\pdfoutput=1

\documentclass[11pt]{article}

\usepackage[final]{acl}

\newcommand{\sharedthanks}{Equal contribution. Work done during internship at Meituan.}

\usepackage{times}
\usepackage{latexsym}
\usepackage{graphicx}  
\usepackage{caption}

\usepackage[T1]{fontenc}

\usepackage[utf8]{inputenc}

\usepackage{microtype}

\usepackage{inconsolata}

\usepackage{booktabs}
\usepackage{multirow}
\usepackage{svg}
\usepackage{amsmath}

\usepackage{natbib}

\title{SCoder: Iterative Self-Distillation for Bootstrapping    \\Small-Scale Data Synthesizers to Empower Code LLMs}

\author{
 \textbf{Xinyu Zhang\textsuperscript{1, \thanks{\sharedthanks}}},
  \textbf{Changzhi Zhou\textsuperscript{1,  \footnotemark[1]}},
 \textbf{Linmei Hu\textsuperscript{1,\thanks{Corresponding author.}}},
 \textbf{Luhao Zhang\textsuperscript{1}},\\
 \textbf{Xiancai Chen\textsuperscript{2}},
 \textbf{Haomin Fu\textsuperscript{3}},
 \textbf{Yang Yang\textsuperscript{3}},
 \textbf{Mengdi Zhang\textsuperscript{3}}
\\
 \textsuperscript{1} School of Computer Science, Beijing Institute of Technology,\\
 \textsuperscript{2} School of Computer Science, Peking University
 \textsuperscript{3} Meituan
\\
\texttt{\{xyzhang0105\}@gmail.com,}  \texttt{\{zhou\_changzhi97, hulinmei\}@bit.edu.cn} 
}

\begin{document}
\maketitle
\begin{abstract}
Existing code large language models (LLMs)  often rely on large-scale instruction data distilled from proprietary  LLMs for fine-tuning, which typically incurs high costs. In this paper, we explore the potential of small-scale open-source LLMs (e.g., 7B) as synthesizers for high-quality code instruction data construction. We first observe that the data synthesis capability of small-scale LLMs can be enhanced by training on a few superior data synthesis samples from proprietary LLMs. Building on this, we propose a novel iterative self-distillation approach to bootstrap small-scale LLMs, transforming them into powerful synthesizers that reduce reliance on proprietary LLMs and minimize costs. Concretely, in each iteration, to obtain diverse and high-quality self-distilled data, we design multi-checkpoint sampling and multi-aspect scoring strategies for initial data selection. Furthermore, to identify the most influential samples, we introduce a gradient-based influence estimation method for final data filtering. Based on the code instruction datasets from the small-scale synthesizers, we develop SCoder, a family of code generation models fine-tuned from DeepSeek-Coder. SCoder models achieve state-of-the-art code generation capabilities, demonstrating the effectiveness of our method.

\end{abstract}

\section{Introduction}

Code generation has long been a central challenge in computer science and has attracted wide attention from the research community. Recent advancements in code
large language models (LLMs) \cite{DBLP:journals/corr/abs-2107-03374, DBLP:journals/corr/abs-2203-07814, DBLP:journals/tmlr/LiAZMKMMALCLZZW23, DBLP:journals/jmlr/ChowdheryNDBMRBCSGSSTMRBTSPRDHPBAI23, DBLP:journals/corr/abs-2308-12950, DBLP:journals/corr/abs-2402-19173} have led to significant breakthroughs. These models can generate code that closely aligns with user intent and are increasingly being widely adopted.

Typically, instruction tuning on base models (e.g., DeepSeek-Coder-Base) is a crucial step in developing high-performance code LLMs.
Therefore, extensive research on code LLMs focuses on constructing high-quality instruction data.  A common approach  involves distilling knowledge from proprietary LLMs. For instance, Code Alpaca \cite{codealpaca} and WizardCoder \cite{DBLP:conf/iclr/LuoX0SGHT0LJ24} are fine-tuned with instruction data distilled from GPT-3.5, using Self-Instruct \cite{DBLP:conf/acl/WangKMLSKH23} and Evol-Instruct \cite{DBLP:conf/iclr/XuSZG0FTLJ24}, respectively. Additionally, MagicoderS \cite{DBLP:conf/icml/0003W0D024} is fine-tuned on data distilled from both GPT-3.5 and GPT-4, using OSS-Instruct to generate coding problems and solutions based on the given code snippets. While these methods have proven effective, they all suffer from the cost-intensive issue caused by the distillation of large-scale instruction data  from the proprietary LLMs like GPT-3.5 and GPT-4.

\begin{figure}[t!] 
    \centering 
    \includegraphics[width=0.49 \textwidth]{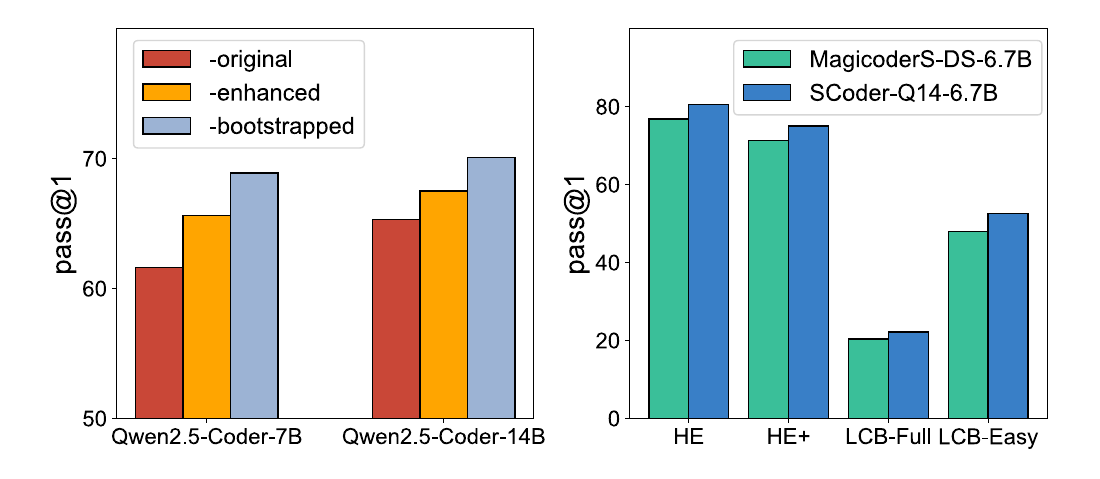} 
    \caption{\textbf{Left}: The performance of code generation models on HumanEval using data provided by different synthesizers (Qwen2.5-Coder-7B or -14B). \textbf{Right}: The performance of our SCoder and the baseline. SCoder uses 60K instruction data generated by a small-scale synthesizer, and the baseline uses 75K instruction data generated by proprietary LLMs. All code generation models are fine-tuned from DeepSeek-Coder-6.7B-Base.}
    \label{intro_fig} 
\end{figure}

In this paper, we explore the potential of relatively small-scale (7B, 8B, and 14B) open-source LLMs as synthesizers for code instruction data construction. Previous works have shown that small LLMs can assist in pre-training data synthesis for non-code domains \cite{yang2024qwen2}. However, instruction data typically takes a different form from pre-training data and requires higher quality standards \cite{wang2025epicoder}. To validate the feasibility of small LLMs in synthesizing code instruction data, we conduct a preliminary experiment. First, we  use small-scale LLMs as original synthesizers and further train them on a limited set of proprietary LLM-distilled samples as enhanced synthesizers. Then, we fine-tune code generation models using data provided by them. The results on the left of Figure~\ref{intro_fig} show that the instruction data provided by the enhanced synthesizer outperforms that of the original, highlighting that a few superior samples can unleash the data synthesis potential of small models. However, distilling more proprietary samples to further improve the synthesis capability of small synthesizers would again trigger the cost-intensive issue. Therefore, a crucial question arises: \textbf{\textit{Can we continuously improve the data synthesis capability of small-scale synthesizers without relying on proprietary LLMs' samples?}}

To address this, we propose a {progressive self-distillation method that iteratively bootstraps} the code instruction data synthesis capability of small-scale LLMs. Specifically, starting with an enhanced synthesizer, we employ a two-step approach in each iteration to obtain high-quality self-distilled data synthesis samples for further training. First, we design \emph{multi-checkpoint sampling} and \emph{multi-aspect scoring} strategies to obtain diverse and reliable self-distilled samples. Then, we introduce a \emph{gradient-based influence estimation} method to further select the influential ones by {comparing the gradients induced by self-distilled samples with those induced by superior samples from proprietary LLMs.}
We validate our method on  small-scale LLMs like Qwen2.5-Coder-7B/14B-Ins \cite{DBLP:journals/corr/abs-2409-12186}, improving their data synthesis capabilities as shown in the left of Figure \ref{intro_fig},
 and transforming them into powerful data synthesizers.

Based on the code instruction datasets provided by our small-scale synthesizers, we develop SCoder, a {family of} code generation models fine-tuned from DeepSeek-Coder-6.7B-Base~\cite{DBLP:journals/corr/abs-2401-14196}. Experimental results on HumanEval (+) \cite{DBLP:journals/corr/abs-2107-03374, DBLP:conf/nips/LiuXW023}, MBPP (+) \cite{DBLP:journals/corr/abs-2108-07732}, LiveCodeBench \cite{DBLP:journals/corr/abs-2403-07974}, and BigCodeBench \cite{DBLP:journals/corr/abs-2406-15877} show that SCoder outperforms or matches state-of-the-art code LLMs that use the instruction data from proprietary LLMs.
Overall, our contributions can be summarized as follows:

\begin{itemize}
    \item {We propose a novel iterative self-distillation approach that transforms small-scale LLMs into effective synthesizers of code instruction data. Using the instruction data generated by these synthesizers, we train a family of code generation models (SCoder), which achieve performance comparable to that of models relying on proprietary LLM-distilled data.}
    \item {To obtain diverse and high-quality self-distilled data, we design multi-checkpoint sampling and multi-aspect scoring strategies for initial data selection. To further identify the most influential samples, we introduce a gradient-based influence estimation method for final data filtering.}
    \item We fine-tune the code generation models (SCoder) based on the datasets generated by our small-scale synthesizers. Experimental results on multiple benchmarks show the effectiveness of our method.
\end{itemize}

\section{Related Work}

\subsection{Code Large Language Models}

In recent years, large language models have garnered unprecedented attention~\cite{zhang2024effective, azaria2024chat, chen2024large, zhou2025refinecoder}, and LLM-driven code generation has achieved remarkable progress. Prominent closed-source models such as Codex~\cite{DBLP:journals/corr/abs-2107-03374}, GPT-4~\cite{DBLP:journals/corr/abs-2303-08774}, PaLM~\cite{DBLP:journals/jmlr/ChowdheryNDBMRBCSGSSTMRBTSPRDHPBAI23}, and Gemini~\cite{DBLP:journals/corr/abs-2312-11805} have shown impressive performance across various code generation benchmarks. Meanwhile, open-source models like CodeGen~\cite{DBLP:conf/iclr/NijkampPHTWZSX23}, CodeGeeX~\cite{DBLP:conf/kdd/ZhengXZDWXSW0LS23}, StarCoder~\cite{DBLP:journals/tmlr/LiAZMKMMALCLZZW23}, CodeLlama~\cite{DBLP:journals/corr/abs-2308-12950}, DeepSeek-Coder~\cite{DBLP:journals/corr/abs-2401-14196}, and CodeQwen~\cite{DBLP:journals/corr/abs-2409-12186} have also made substantial contributions. These models not only enhance code generation capabilities but also promote more efficient and automated software development.

Typically, such models are developed through continual pre-training~\cite{DBLP:journals/corr/abs-2308-12950}, followed by supervised fine-tuning (SFT)~\cite{DBLP:journals/corr/abs-2312-14187}. While pre-training utilizes large-scale, unannotated code corpora, SFT relies on high-quality labeled instruction data, whose construction remains a key challenge~\cite{DBLP:conf/acl/DingQZLLCX0LJ24}.

\subsection{Code Instruction Data Synthesis}

Creating diverse and complex code instruction data is challenging and requires domain expertise. While human-written datasets used in OctoPack~\cite{DBLP:conf/iclr/MuennighoffLZZH24} and PIE~\cite{DBLP:conf/iclr/ShypulaMZ0GYHNR24} are effective, they are labor-intensive and hard to scale. To address this, many recent works leverage powerful proprietary LLMs for automatic instruction generation. For example, Code Alpaca~\cite{codealpaca} adopts Self-Instruct~\cite{DBLP:conf/acl/WangKMLSKH23}, WizardCoder~\cite{DBLP:conf/iclr/LuoX0SGHT0LJ24} uses Evol-Instruct~\cite{DBLP:conf/iclr/XuSZG0FTLJ24}, and Magicoder~\cite{DBLP:conf/icml/0003W0D024} utilizes OSS-Instruct to create realistic, diverse programming tasks from open-source code. Similarly, WaveCoder~\cite{DBLP:journals/corr/abs-2312-14187} introduces a generator-discriminator framework, while OpenCodeInterpreter~\cite{DBLP:conf/acl/ZhengZSLLFCY24} leverages user-LLM-compiler interactions to synthesize multi-turn instruction data. Despite their effectiveness, these approaches often depend on costly proprietary models~\cite{DBLP:journals/corr/abs-2407-05700}. In this work, we explore using small-scale open-source LLMs to generate high-quality code instruction data more cost-effectively, reducing reliance on expensive proprietary models while maintaining strong performance.

\section{Methodology}

\subsection{Overview}

In this work, we aim to train a set of small-scale code instruction data synthesis models, named synthesizers, capable of generating high-quality code instruction data, i.e., the code problem-solution pair $(q, s)$ given an open-source code snippet $c$ and an instruction synthesis prompt $p$. 
To achieve this, we first construct a clean and noise-free code snippet pool $\mathcal{C} = \{c_i\}$, following the data pre-processing pipeline of StarCoder2~\cite{DBLP:journals/corr/abs-2402-19173}.
Next, we distill a limited number of instruction data synthesis samples, denoted as 
$\mathcal{D}_{p} = \{ (p, c_i^{p}, q_i^{p}, s_i^{p}) \}$, 
from proprietary LLMs to obtain enhanced synthesizers. Finally, we propose an iterative bootstrap approach to continuously train the synthesizers using self-distilled data, denoted as
$\mathcal{D}_{s} = \{ (p, c_i^{{s}}, q_i^{{s}}, s_i^{{s}}) \}$. The prompt $p$ and more details of the code snippet pool $\mathcal{C}$ are provided in Appendix~\ref{magicoder_prompt} and ~\ref{code seed gather}, respectively.

\begin{table}[t!]
\small
  \centering
    \begin{tabular}{lcc}
        \toprule
          \textbf{Synthesizer} & \textbf{HumanEval} & \textbf{MBPP} \\
        \midrule
        Llama3.1-8B-Ins & 60.4  & 64.7  \\
    \ \ \ \ \textit{+Enhanced} & 64.2  & 69.3  \\
    \midrule
    Qwen2.5-Coder-7B-Ins & 61.6  & 70.8  \\
    \ \ \ \ \textit{+Enhanced} & 65.6  & 72.1  \\
    \midrule
    Qwen2.5-Coder-14B-Ins & 65.3  & 73.7  \\
    \ \ \ \ \textit{+Enhanced} & 67.5  & 75.8  \\
        \bottomrule
    \end{tabular}%
  \caption{The performance of the code generation model fine-tuned on 40K code instruction data provided by different synthesizers.}
  \label{pre_exp}%
\end{table}%

\subsection{Preliminary Study}
We conduct a preliminary study to validate whether small LLMs can acquire a certain level of data synthesis capability by distilling a limited number of proprietary LLM samples.
To obtain proprietary samples with sufficient knowledge coverage, we adopt a classification-based diversified code snippet sampling technique. Specifically, we employ 10 pre-defined task categories and calculate the similarity between each code snippet and the task category descriptions with the help of a state-of-the-art embedding model INSTRUCTOR~\cite{DBLP:conf/acl/SuSKWHOYSZ023}. Based on the embedding similarity, each code snippet is assigned to its most relevant task category. We then randomly sample 1K code snippets from each category to ensure sufficient knowledge diversity. Finally, these selected code snippets are used to prompt proprietary LLMs generating code instruction data synthesis samples $\mathcal{D}_{p} = \{ (p, c_i^{p}, q_i^{p}, s_i^{p}) \}$, where $(p, c_i^{p})$ denotes input and $(q_i^{p}, s_i^{p})$ denotes output.

We use Llama3.1-8B-Ins and Qwen2.5-Coder-7B/14B-Ins as the original synthesizers and train them on $\mathcal{D}_{p}$ to obtain enhanced synthesizers. Based on code instruction data provided by these synthesizers, we fine-tune DeepSeek-Coder-6.7B-Base as the code generation model. The results are shown in Table~\ref{pre_exp}, the enhanced synthesizers exhibit a significant improvement in data synthesis capability, even with only 10K proprietary LLM samples. This demonstrates the strong potential of small models for code instruction data synthesis.

\begin{figure*}[t!] 
    \centering 
    \includegraphics[width=1\textwidth]{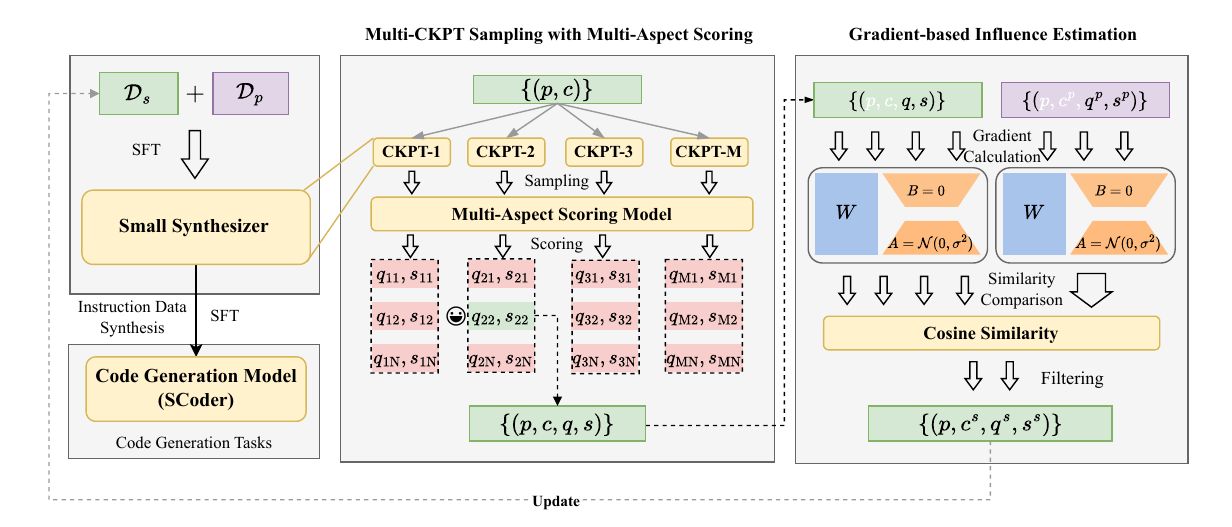} 
    \caption{\textbf{Overview of our iterative self-distillation bootstrap method.} In each iteration, we sample outputs from multiple checkpoints and evaluate them with a multi-aspect scorer for diversity and reliability. We then use a gradient-based influence estimation method to select the most influential samples, which is done by evaluating the gradient similarity between the self-distilled and proprietary LLM-distilled code instruction data.} 
    \label{SCoder_fig} 
\end{figure*}

\subsection{Bootstrapping with Iterative Self-Distillation}

To further boost small LLMs for synthesizing higher-quality code instruction data without distilling additional proprietary LLM samples, in this section, we propose an effective bootstrap method based on iterative self-distillation. Specifically, we start with the mentioned enhanced synthesizers, considering this as the 0-th iteration of the bootstrap. Then, in each iteration, we first collect diverse and reliable self-distilled data synthesis samples by multi-checkpoint sampling and multi-aspect scoring strategies. These samples are generated by the synthesizers from the previous iteration. Next, to further identify the most influential samples, we introduce a gradient-based influence estimation method, which quantifies each sample's influence by computing its gradient similarity with proprietary LLM samples. Finally, these high-quality samples are used to train the synthesizer itself, enhancing its ability to generate code instruction data. The overview of our method is illustrated in Figure \ref{SCoder_fig}, and a detailed theoretical analysis of the iterative self-distillation is provided in Appendix \ref{sec:theoretical_analysis}.

\paragraph{Multi-Checkpoint Sampling with Multi-Aspect Scoring.}
As our approach iteratively trains on self-distilled data synthesis samples, ensuring their quality and diversity is essential. 
Therefore, we first develop a multi-checkpoint sampling strategy. Specifically, given the synthesis prompt $p$ and a code snippet $c$, we obtain $M \times N$ diverse problem-solution pairs $\{ (q_{ij}, s_{ij}) \}$ by sampling $N$ times from $M$ checkpoints of synthesizers, where $i \in [1,M]$ and $j \in [1,N]$. Compared to the strategy Best-of-N~\cite{stiennon2022learningsummarizehumanfeedback}, which selects candidates from a single checkpoint, our approach expands the search space and improves both the reliability and diversity of the selected data.

Next, to rank and select the best candidate pair corresponding to the code snippet, we introduce a multi-aspect scoring model, namely scorer. Given a candidate pair $(q_{ij}, s_{ij})$, the scorer
evaluates it across $Z$ aspects, producing a feature vector 
$\mathbf{x}_{ij} = \{x_{ij}^{z}\}$
, where $x_{ij}^{z} \in [0,9]$ represents the integer score in the $z$-th aspect, such as problem-solution consistency \footnote{The prompt for the multi-aspect scorer are provided in Appendix \ref{scorer}.}. Furthermore, considering that different aspects are independent and integer-based scores provide only a hard signal that lacks granularity for distinguishing data quality, we propose a weighted scoring aggregation method, which  assigns each aspect
a weight $w^{z}$ and computes the final aggregated real-valued score $Score_{ij}$ as:
\begin{equation}
    Score_{ij} = \sum_{z=1}^{Z} w^{z} x_{ij}^{z}.
\end{equation}
To determine the optimal weight vector $\mathbf{w} = \{w^{z}\}$, we conduct $K$ experiments based on the instruction data generated by synthesizers.
For each experiment, we compute the average multi-aspect scores $\bar{\mathbf{x}}_k$ of the instruction data and use the data to fine-tune DeepSeek-Coder-6.7B-Base. The fine-tuned model is then evaluated on an out-of-distribution (OOD) test set to obtain the corresponding performance score $y_k$. Given the data $\{(\bar{\mathbf{x}}_k, y_k)\}$, we estimate $\mathbf{w}$ by solving the following ridge regression problem:
\begin{equation}
    \mathbf{w} = \arg\min_{\mathbf{w}} \sum_{k=1}^{K} (y_{k} - \mathbf{w} \cdot \bar{\mathbf{x}}_{k})^2 + \lambda \|\mathbf{w}\|^2,
\end{equation}
where $\lambda$ is a regularization term to prevent overfitting, and the learned weights indicate the relative importance of each scoring aspect in determining the effectiveness of instruction data.

\paragraph{Gradient-based Influence Estimation}
While multi-checkpoint sampling with multi-aspect scoring ensures diversity and reliability, the influence of each selected self-distilled sample on the fine-tuning of the base model can vary. Inspired by previous works \cite{DBLP:conf/nips/PruthiLKS20, DBLP:conf/icml/XiaMGA024}, we introduce a gradient-based influence estimation method to further identify the most valuable samples by estimating the fine-tuning influence of the code instruction data they contain.

Concretely, based on the influence formulation \cite{DBLP:conf/nips/PruthiLKS20}, the influence of a self-distilled code instruction data $d=(q, s)$ on the prediction of a test instance $t$ in a base model parameterized by $\theta$ can be estimated by computing the similarity between their gradients: 
\begin{equation}
    \mathrm{Inf}(d,t)\propto\mathrm{Sim}(\nabla l(d,\theta),\nabla l(t,\theta)).
\end{equation}
However, code generation tasks are inherently broad and diverse, and some of them may lack well-established benchmarks. To address this, we instead estimate the influence of $d$ by computing its gradient similarity to the code instruction data $\{d^p=(q^p, s^p)\}$ from proprietary LLM samples $\mathcal{D}_{p}$. The idea is that proprietary LLMs (e.g., GPT-4o) have undergone extensive optimization through various strategies, making their distilled instruction data highly effective in improving model performance across diverse tasks.

Specifically, inspired by previous work \cite{DBLP:conf/icml/XiaMGA024}, we first train an LLM-based reference model on the proprietary instruction data $\{d^p=(q^p, s^p)\}$ using LoRA \cite{DBLP:conf/iclr/HuSWALWWC22}, which allows for low-rank adaptation, significantly reducing trainable parameters and ensuring the efficiency for the following gradient computations.
We then compute the gradient of each self-distilled instruction data $d$ with respect to the LoRA parameters $\theta_{{lora}}$, denoted as $\nabla l_{{ref}}(d, \theta_{{lora}})$. To further improve efficiency, following prior work \cite{DBLP:conf/icml/ParkGILM23}, we apply a projection matrix initialized with a Rademacher distribution to reduce gradient dimensionality, resulting in 
    $\hat{\nabla}l_{{ref}}(d, \theta_{{lora}})$.
According to the Johnson-Lindenstrauss Lemmas \cite{johnson1984extensions}, this transformation can preserve gradient distances while ensuring the usefulness of lower-dimensional features. Similarly, we compute the projected gradients for each proprietary instruction data $d^p$, denoted as $\hat{\nabla}l_{{ref}}(d^p, \theta_{{lora}})$. Finally, we approximate the influence of $d$  by calculating its cosine similarity to the average gradient of $\{d^p\}$:
\begin{equation}
    \begin{split}
        V(d) = \mathsf{Cosine} \Bigg( 
        \hat{\nabla}l_{{ref}}(d, \theta_{{lora}}),  \\
        \frac{1}{N_{{p}}} \sum_{i=1}^{N_{{p}}} 
        \hat{\nabla}l_{{ref}}(d_{i}^p, \theta_{{lora}}) 
        \Bigg),
    \end{split}
\end{equation}
where $N_{{p}}$ is the size of $\{d^p\}$. Eventually, the data samples with the highest influence will be selected and used for training.

\setlength{\tabcolsep}{5pt}
\begin{table*}[t!]
\small
  \centering
  \label{bootstrap}
    \begin{tabular}{lc|cccc}
    \toprule[1pt]
    \textbf{Synthesizer}  & \textbf{Data Size} & \textbf{HumanEval} & \textbf{MBPP}  & \textbf{LiveCodeBench} & \textbf{BigCodeBench} \\
    \midrule
    \multicolumn{6}{c}{\textbf{DeepSeek-Coder-6.7B-Base}} \\
    \midrule
    None & 0 & 47.6\textsuperscript{†} & 72.0\textsuperscript{†} & 16.2\textsuperscript{†} & 41.8\textsuperscript{†} \\
    \midrule
    \multicolumn{6}{c}{\textbf{Fine-Tuning DeepSeek-Coder-6.7B-Base on 40K Synthesized Data}} \\
    \midrule
    Llama3.1-8B-Instruct & 0 &  60.4  & 64.7  & 16.5  & 42.1  \\
    \ \ \ \ \textit{+Enhanced} & 10K & 64.2  & 69.3  & 17.3  & 42.8 \\
    \ \ \ \ \textit{+1 Iter} & 20K & 65.5  & 71.1  & 17.4  & 43.1 \\
    \ \ \ \ \textit{+2 iter} & 40K & \textbf{67.4}  & \textbf{73.4}  & \textbf{17.8}  & \textbf{43.5} \\
    \midrule
    Qwen2.5-Coder-7B-Instruct & 0 & 61.6  & 70.8  & 17.0  & 42.7  \\
    \ \ \ \ \textit{+Enhanced} & 10K & 65.6  & 72.1  & 18.2  & 43.8 \\
    \ \ \ \ \textit{+1 Iter} & 20K & 66.3  & 72.9  & 18.4  & 44.1 \\
    \ \ \ \ \textit{+2 iter}  & 40K & \textbf{68.9} & \textbf{74.7} & \textbf{18.9} & \textbf{44.7} \\
    \midrule
    Qwen2.5-Coder-14B-Instruct &  0 &65.3  & 73.7  & 18.7  & 43.2 \\
    \ \ \ \ \textit{+Enhanced} &  10K &67.5  & 75.8  & 19.4  & 44.5 \\
    \ \ \ \ \textit{+1 Iter} &  20K &68.4  & 76.3  & 19.3  & 45.1 \\
    \ \ \ \ \textit{+2 iter} &  40K &\textbf{70.1}  & \textbf{76.5}  & \textbf{19.7}  & \textbf{45.9} \\
    \bottomrule[1pt]
    \end{tabular}
    \caption{Performance of code generation models (target models) built on instruction data generated by small synthesizers on HumanEval, MBPP, LiveCodeBench (Full), and BigCodeBench (Complete-Full). Data size refers to the amount of data used to train the synthesizer. † denotes results from the benchmark leaderboards.}
  \label{bootstrap_table}
\end{table*}


\begin{table*}[t!]
\small
  \centering
    \begin{tabular}{l|llllllllll}
    \toprule
    \multirow{2}[0]{*}{\textbf{Models}} & \multicolumn{2}{c}{\textbf{HumanEval}} & \multicolumn{2}{c}{\textbf{MBPP}} & \multicolumn{2}{c}{\textbf{LiveCodeBench}} & \multicolumn{2}{c}{\textbf{BCB (Comp)}} & \multicolumn{2}{c}{\textbf{BCB (Inst)}} \\
    \cmidrule(lr){2-3} \cmidrule(lr){4-5} \cmidrule(lr){6-7} \cmidrule(lr){8-9} \cmidrule(lr){10-11}
          & Base  & Plus  & Base  & Plus  & Full  & Easy  & Full  & Hard  & Full  & Hard \\
    \midrule
    \multicolumn{11}{c}{\textbf{Proprietary Models}} \\
    \midrule
    GPT-4-Turbo-20240409 & 90.2\textsuperscript{†} & 86.6\textsuperscript{†} & 85.7\textsuperscript{‡} & 73.3\textsuperscript{‡} & 42.0\textsuperscript{†} & 82.4\textsuperscript{†} & 58.2\textsuperscript{†} & 35.1\textsuperscript{†} & 48.2\textsuperscript{†} & 32.1\textsuperscript{†} \\
    GPT-o1-Preview-20240912 & 96.3\textsuperscript{†} & 89.0\textsuperscript{†} & 95.5\textsuperscript{†} & 80.2\textsuperscript{†} & 58.5\textsuperscript{†} & 94.1\textsuperscript{†} & / & 34.5\textsuperscript{†} & / & 23.0\textsuperscript{†} \\
    \midrule
    \multicolumn{11}{c}{\textbf{DeepSeek-Coder-6.7B-Base}} \\
    \midrule
    DeepSeek-Coder-6.7B-Base & 47.6\textsuperscript{†} & 39.6\textsuperscript{†} & 72.0\textsuperscript{†} & 58.7\textsuperscript{†} & 16.2\textsuperscript{†} & 38.7\textsuperscript{†} & 41.8\textsuperscript{†} & 13.5\textsuperscript{†} & / & / \\
    \midrule
    \multicolumn{11}{c}{\textbf{Fine-Tuned Models based on DeepSeek-Coder-6.7B-Base}} \\
    \midrule
    DeepSeek-Coder-6.7B-Instruct & 74.4\textsuperscript{†} & 71.3\textsuperscript{†} & 74.9\textsuperscript{†} & 65.6\textsuperscript{†} & 19.8\textsuperscript{†} & 45.8\textsuperscript{†} & 43.8\textsuperscript{†} & 15.5\textsuperscript{†} & 35.5\textsuperscript{†} & 10.1\textsuperscript{†} \\
    WaveCoder-Ultra-6.7B & 75.0\textsuperscript{†} & 69.5\textsuperscript{†} & 74.9\textsuperscript{†} & 63.5\textsuperscript{†} & 19.7  & 46.8  & 43.7\textsuperscript{†} & \textbf{16.9\textsuperscript{†}} & 33.9\textsuperscript{†} & 12.8\textsuperscript{†} \\
    MagicoderS-DS-6.7B & 76.8\textsuperscript{†} & 71.3\textsuperscript{†} & \underline{79.4}\textsuperscript{†} & \underline{69.0}\textsuperscript{†} & 20.4  & 47.9  & \underline{47.6}\textsuperscript{†} & 12.8\textsuperscript{†} & 36.2\textsuperscript{†} & 13.5\textsuperscript{†} \\
    OpenCodeInterpreter-DS-6.7B & 77.4\textsuperscript{†} & 71.3\textsuperscript{†} & 76.5\textsuperscript{†} & 66.4\textsuperscript{†} & 18.9  & 46.6  & 44.6\textsuperscript{†} & \textbf{16.9\textsuperscript{†}} & 37.1\textsuperscript{†} & 13.5\textsuperscript{†} \\
    AlchemistCoder-DS-6.7B & \underline{79.9}\textsuperscript{‡} & \underline{75.6}\textsuperscript{‡} & 77.0\textsuperscript{‡} & 60.2\textsuperscript{‡} & 17.4  & 44.7  & 42.5  & 14.2  & 33.5  & 13.2  \\
    InverseCoder-DS-6.7B & \underline{79.9}\textsuperscript{‡} & \textbf{76.8\textsuperscript{‡}} & 78.6\textsuperscript{‡} & \underline{69.0}\textsuperscript{‡} & 20.3  & 46.6  & 45.7  & 14.9  & 35.4  & 9.5  \\
    WizardCoder-GPT-4-6.7B & 77.4  & 73.8  & 75.4  & 64.8  & 21.0  & 49.6  & 45.1  & 15.5  & 37.3  & 10.8  \\
    \midrule
    SCoder-L-DS-6.7B & 78.2  & 73.8  & 77.6  & 65.4  & 21.1  & 51.7  & 46.2  & 15.1  & 37.9  & 13.4  \\
    SCoder-Q7-DS-6.7B & 78.7  & 74.3  & 79.1  & 66.5  & \underline{21.4}  & \underline{52.2}  & 47.4  & 15.5  & \underline{38.6}  & \underline{14.5}  \\
    SCoder-Q14-DS-6.7B & \textbf{80.5} & 75.0  & \textbf{81.0} & \textbf{69.3} & \textbf{22.2} & \textbf{52.6} & \textbf{49.2}  & \underline{16.2}  & \textbf{40.6} & \textbf{16.9} \\
    
    \bottomrule
    \end{tabular}%
    \caption{Performance comparison of different models on multiple code generation benchmarks. Three SCoder models are fine-tuned using data generated by our small synthesizers, where L, Q7, and Q14 denote three different synthesizers after two iterations of bootstrap. BCB, Comp, and Inst denote BigCodeBench, Complete, and Instruct. ‡ denotes results from the InverseCoder work~\cite{DBLP:journals/corr/abs-2407-05700}. The best results are in \textbf{bold} and the second-best results are \underline{underlined}. }
  \label{all_results}%
\end{table*}

\begin{table*}[t!]
  \centering
  \small
    \begin{tabular}{lcccc}
    \toprule
    \textbf{Models} & \textbf{HumanEval} & \textbf{MBPP}  & \textbf{LiveCodeBench} & \textbf{BigCodeBench} \\
    \midrule
    \midrule
    SCoder-Q7-DS-6.7B & \textbf{78.7}  & \textbf{79.1}  & \textbf{21.4}  & \textbf{47.4}  \\
    w/o multi-checkpoint sampling & 74.9  & 73.8  & 18.7  & 44.3  \\
    w/o multi-aspect scoring & 72.3  & \underline{76.7}  & \underline{19.9}  & \underline{45.5}  \\
    w/o gradient-based influence estimation & \underline{75.1}  & 74.4  & 18.2  & 43.2  \\
    \midrule
    \midrule
    SCoder-Q14-DS-6.7B & \textbf{80.5}  & \textbf{81.0}  & \textbf{22.2}  & \textbf{49.2}  \\
    w/o multi-checkpoint sampling & 75.6  & 74.4  & 20.4  & \underline{46.3}  \\
    w/o multi-aspect scoring & 74.9  & \underline{75.8}  & \underline{20.8}  & 45.1  \\
    w/o gradient-based influence estimation & \underline{76.1}  & 74.9  & 20.0  & 44.8  \\
    \bottomrule
    \end{tabular}%
    \caption{Ablation study on HumanEval, MBPP, LiveCodeBench (Full), and BigCodeBench (Complete-Full). The best results are in \textbf{bold} and the second-best results are \underline{underlined}.}
  \label{ablation_study}%
\end{table*}%

\section{Experiments}
\subsection{Benchmarks}
We evaluate model performance using the pass@1 metric on several standard benchmarks: HumanEval \cite{DBLP:journals/corr/abs-2107-03374}, MBPP \cite{DBLP:journals/corr/abs-2108-07732} (along with their EvalPlus \cite{DBLP:conf/nips/LiuXW023} versions), LiveCodeBench (V4) \cite{DBLP:journals/corr/abs-2403-07974}, and BigCodeBench \cite{DBLP:journals/corr/abs-2406-15877}. Evaluation strictly follows each benchmark's official settings and prompts.

\subsection{Baselines}
We compare SCoder with several powerful baselines, including two proprietary models: GPT-4-Turbo-20240409 \cite{gpt4-blog} and GPT-o1-Preview-20240912 \cite{o1-blog}, as well as seven open-source models built on DeepSeek-Coder-6.7B-Base~\cite{DBLP:journals/corr/abs-2401-14196}: DeepSeek-Coder-6.7B-Instruct, WaveCoder-Ultra-6.7B \cite{DBLP:journals/corr/abs-2312-14187}, MagicoderS-DS-6.7B \cite{DBLP:conf/icml/0003W0D024}, OpenCodeInterpreter-DS-6.7B \cite{DBLP:conf/acl/ZhengZSLLFCY24}, AlchemistCoder-DS-6.7B~\cite{song2024alchemistcoder}, InverseCoder-DS-6.7B~\cite{DBLP:journals/corr/abs-2407-05700}, and WizardCoder-GPT-4-6.7B~\cite{DBLP:conf/iclr/LuoX0SGHT0LJ24}.

\subsection{Implementation Details} 

We provide a simplified version of the implementation details here; a more detailed version can be found in Appendix \ref{imple_detail}.

\begin{table}[t!]
  \centering
  \resizebox{0.48\textwidth}{!}{
    \begin{tabular}{ccc}
    \toprule
    \multicolumn{1}{c}{\textbf{Model}} & \multicolumn{1}{c}{\textbf{Common Data}} & \multicolumn{1}{c}{\textbf{Specific Data}} \\
    \midrule
    WizardCoder-GPT-4 & \multirow{6}{*}{110K (GPT-4)} & 0K \\
    WaveCoder-Ultra &       & 20K (GPT-4) \\
    MagicoderS &       & 75K (GPT-3.5) \\
    AlchemistCoder &       & >80K (GPT-3.5) \\
    InverseCoder &       & 90K (self-generated) \\
    SCoder (ours) &       & 60K (small model-generated) \\
    \bottomrule
    \end{tabular}
  }
  \caption{Comparison of data used by different models. The source of the data is indicated in parentheses.}
  \label{data_used}
  \vspace{-0.1in}
\end{table}

\paragraph{Small-Scale Data Synthesizer.}  
We train Llama3.1-8B-Ins, Qwen2.5-Coder-7B-Ins, and Qwen2.5-Coder-14B-Ins as data synthesizers. Each model is first trained on 10K GPT-4o data $D_p$, then bootstrapped with 20K and 40K self-distilled data $D_s$. Training uses a learning rate of $1 \times 10^{-5}$, global batch size 128, and inference temperature 0.2.

\paragraph{SCoder.}  
For fair comparison, we train DeepSeek-Coder-6.7B-Base on 110K evol-codealpaca-v1 for 2 epochs, then fine-tune it on 60K synthesized data (from small synthesizers) for 3 epochs to obtain SCoder. The 110K data is commonly used in baselines (Table \ref{data_used}). More target model results are in Appendix \ref{moretargetmodel}.

\subsection{Main Results}
As shown in Table \ref{bootstrap_table}, our proposed method significantly enhances the instruction data synthesis capabilities of small models with only two iterations of bootstrap, regardless of their model family or scale. For example, the fine-tuning performance of the 40K data synthesized by Llama3.1-8B-Ins on the base model achieves a 5.0\% improvement on HumanEval and a 5.9\% improvement on MBPP after two iterations of bootstrap. This demonstrates that our approach, leveraging well-designed sampling and filtering strategies, enables small models to acquire self-distilled data synthesis samples with broad diversity, strong reliability,  and high influence. As a result, they progressively evolve into effective data synthesizers while minimizing dependence on proprietary LLM distillation.

Furthermore, Table \ref{all_results} shows that SCoder, trained on data generated by bootstrapped small-scale data synthesizers, outperforms or matches other state-of-the-art open-source baselines across multiple benchmarks. For example, SCoder-Q14-DS-6.7B surpasses the best open-source baselines by 5.9\% and 9.7\% on average in the challenging LiveCodeBench and BigCodeBench, respectively.
Notably, the open-source baselines typically utilize a larger amount of proprietary LLM-distilled instruction data as listed in Table \ref{data_used}, further validating the effectiveness of our method in constructing strong small-scale data synthesizers. A more detailed cost efficiency analysis of our method is provided in Appendix~\ref{appendix:cost}.

\subsection{Ablation Study}
We conduct ablation studies based on SCoder-Q7-DS-6.7B and SCoder-Q14-DS-6.7B. The results presented in Table \ref{ablation_study} demonstrate the importance of our extensive sampling and refined filtering strategies. 

First, without multi-checkpoint sampling (i.e., sampling an equal number of outputs solely from the last checkpoint of the previous iteration), the performance of both code generation models on HumanEval and LiveCodeBench drops by at least 4.8\% and 8.1\%, respectively. This indicates that a limited sampling space reduces the likelihood of obtaining high-quality self-distilled data, thereby hindering the effectiveness of the bootstrap process.  Furthermore, when either multi-aspect scoring or gradient-based influence estimation is removed from the data selection process, the performance on MBPP and BigCodeBench drops by up to 7.5\% and 8.9\%, respectively. This highlights that both strategies are essential for ensuring the reliability and influence of self-distilled data, and removing either significantly impacts the overall effectiveness.

\subsection{Data Scaling}
 
\begin{figure}[t!] 
    \centering 
    \includegraphics[width=0.49 \textwidth]{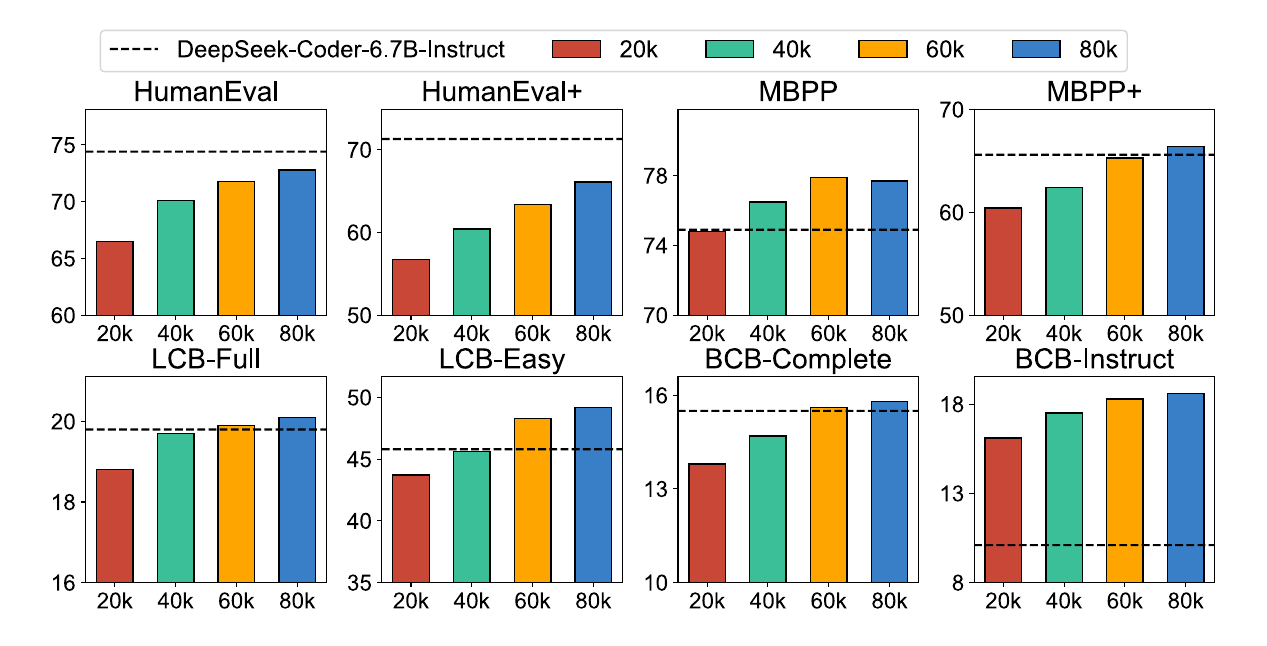} 
    \caption{Impact of data scaling. The dashed lines represent the performance of DeepSeek-Coder-6.7B-Instruct across various benchmarks.} 
    \label{scale_law} 
\end{figure}

To further evaluate the data synthesis quality of small data synthesizers, we investigate the data scaling law using the bootstrapped Qwen2.5-Coder-14B-Ins. As shown in Figure \ref{scale_law}, increasing the data size leads to significant improvements of the code generation model fine-tuned on DeepSeek-Coder-6.7B-Base, surpassing DeepSeek-Coder-6.7B-Instruct on most benchmarks. This further validates the effectiveness of our approach in constructing high-quality small-scale data synthesizers.

\begin{figure}[t!] 
    \centering 
    \includegraphics[width=0.49 \textwidth]{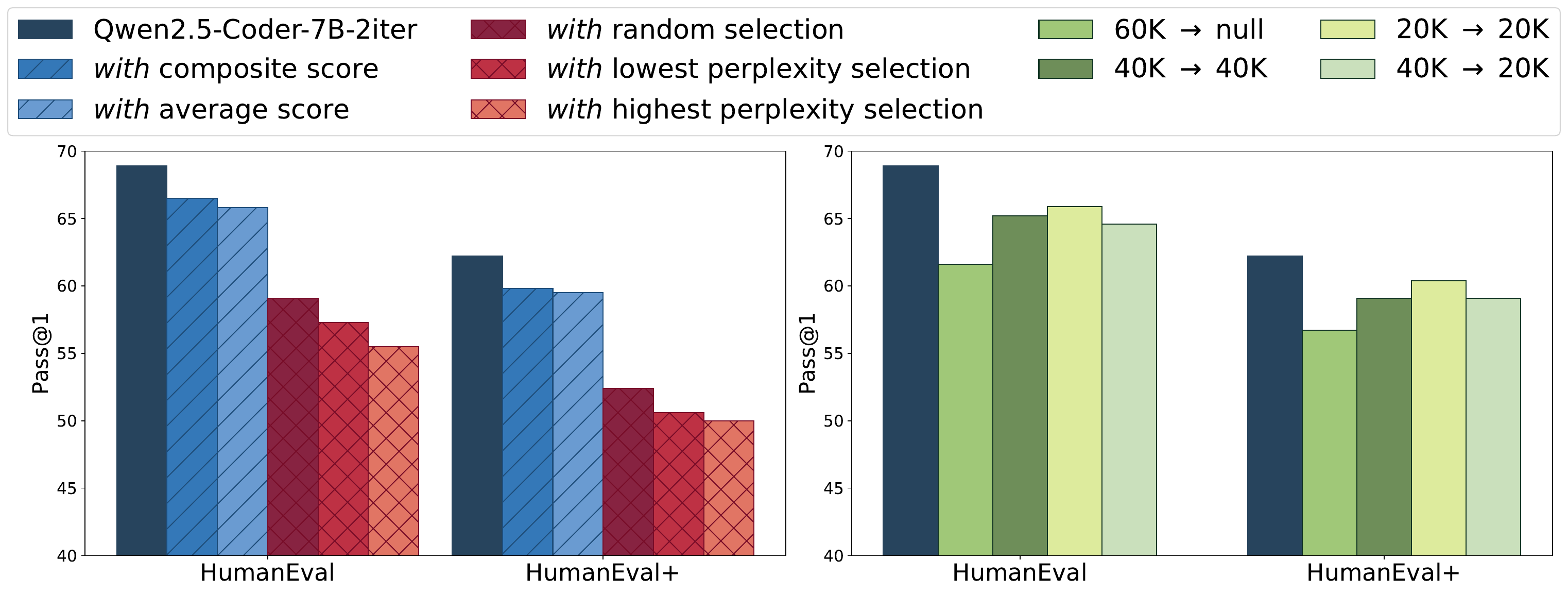} 
    \caption{Comparison of different selection methods
and the number of self-distilled data used in different bootstrap iterations. The y-axis denotes the performance of the code generation models fine-tuned on 40K
synthesized data.} 
    \label{further_dis} 
\end{figure}

\subsection{Further Discussion}

In this section, we provide a more fine-grained analysis of the effectiveness of our method.

First, we compare the impact of different selection strategies during the bootstrap process. As shown on the left of Figure \ref{further_dis}, for multi-aspect scoring, replacing the aggregated score with either the raw composite score from the scorer or the simple average of scores leads to a decline in the synthesizer's data synthesis performance. Moreover, substituting the gradient-based influence estimation with alternative selection methods, such as random selection or lowest/highest perplexity selection, results in an even more substantial performance drop. These findings highlight the effectiveness of our selection strategy in identifying reliable and influential self-distilled samples, thereby ensuring the success of the bootstrap process.

Second, as the synthesizer's capability improves with more bootstrap iterations, we progressively increase the number of self-distilled samples used in training across two iterations (20K $\rightarrow$ 40K). Here, we compare different settings, including removing multi-round iteration (60K $\rightarrow$ Null), progressively decreasing the sample size (40K $\rightarrow$ 20K), increasing the sample size in the first iteration (40K $\rightarrow$ 40K), and decreasing the sample size in the second iteration (20K $\rightarrow$ 20K). As shown on the right of Figure \ref{further_dis}, in all cases, performance declines, indicating that a well-balanced and progressively increasing data schedule plays a crucial role in maximizing the effectiveness of the bootstrap process.

\subsection{Data Quality Analysis}
\begin{figure}[t!] 
    \centering 
    \includegraphics[width=0.3\textwidth]{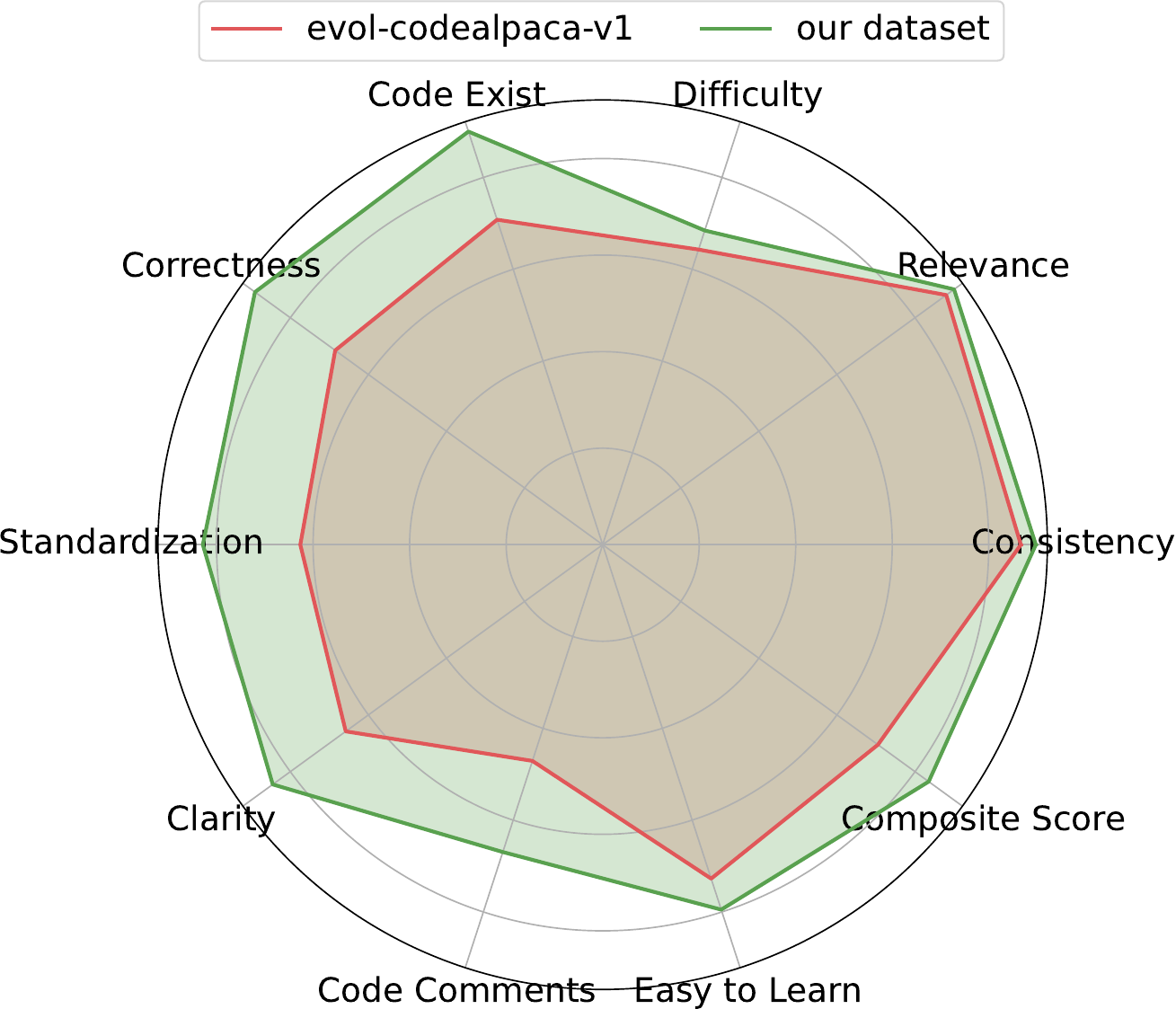} 
    \caption{Quality comparison between the evol-codealpaca-v1 dataset and our synthesized dataset.} 
    \label{data_quality_fig} 
\end{figure}

To further validate the quality of data generated by the synthesizers, we sampled 100 code instruction data from evol-codealpaca-v1 and the bootstrapped Qwen2.5-Coder-14B-Ins, respectively, and used GPT-4o-20240513 and GPT-4-turbo-20240409 to score the data across 10 aspects based on the prompt provided in Appendix \ref{scorer}. The average results, shown in Figure \ref{data_quality_fig}, demonstrate that our synthesized data achieves higher scores across all aspects, further confirming the effectiveness of our method in building high-quality small-scale code instruction synthesizers.

\section{Conclusion}

In this paper, we propose an iterative self-distillation bootstrap method to fully unlock the data synthesis potential of small-scale LLMs, transforming them into powerful code instruction data synthesizers while reducing reliance on proprietary LLMs and minimizing costs. We design multi-checkpoint sampling and multi-aspect scoring strategies to ensure the diversity and reliability of self-distilled samples, followed by a gradient-based influence estimation method to select influential ones for training. We validate our method on Llama3.1-8B-Ins and Qwen2.5-Coder-7B/14B-Ins, demonstrating their effectiveness as data synthesizers. Based on the data generated by these small-scale synthesizers, we introduce SCoder, a family of code generation models that achieves strong performance on HumanEval (+), MBPP (+), LiveCodeBench, and BigCodeBench, showcasing the potential of small models in code instruction data synthesis.

\section{Limitations}

Despite the demonstrated effectiveness of our iterative self-distillation bootstrap method in fully leveraging the code instruction data synthesis capability of small-scale LLMs, certain limitations persist. For example, the current synthesis framework does not incorporate alternative data generation paradigms, such as Self-Instruct \cite{DBLP:conf/acl/WangKMLSKH23} and Evol-Instruct \cite{DBLP:conf/iclr/XuSZG0FTLJ24}, which have shown promise in previous work. Investigating the integration of such approaches constitutes an important direction for future work.

Furthermore, this study limits its empirical validation to the domain of code generation. While the underlying methodology may apply to other domains, several challenges arise. For example, although synthesizers can efficiently generate large-scale code instruction data by leveraging vast amounts of open-source code snippets, achieving efficient data synthesis for other tasks may require additional consideration and tailored design. Therefore, further exploration is needed to fully assess feasibility in other domains, and we plan to present related findings in future work.

\section*{Acknowledgements}

This work was supported by the National Science Foundation of China (No. 62276029), CCF-Zhipu.AI Large Model Fund (No. 202217), Beijing Institute of Technology Research Fund Program for Young Scholars (No.6120220261), and CIPSC-SMP-Zhipu Large Model Cross-Disciplinary Fund.

\bibliography{custom}

\appendix

\newpage
\newpage
\section{Code Snippet Gathering}
\label{code seed gather}
To ensure the validity of our experimental results, we first construct a clean and noise-free code snippet pool that serves as the foundation for code instruction data synthesis. Specifically, inspired by the data preprocessing pipeline of StarCoder2 \cite{DBLP:journals/corr/abs-2402-19173}, we follow the steps below to construct the code snippet pool $\mathcal{C}$ from the Stack V1, a collection of source code in over 300 programming languages.
\begin{itemize}
    \item \textbf{Code Snippet Extraction:} We first extract all Python functions that include docstrings from the Stack V1 dataset. To ensure a high level of diversity while minimizing redundancy, we perform near-deduplication using MinHash, Locality-Sensitive Hashing (LSH), and Jaccard similarity with a threshold of 0.5.
    \item \textbf{Invalid Function Filtering:} We remove any functions that do not contain a return statement or contain syntax errors. Additionally, we supplement the remaining functions with necessary dependency packages and remove functions that import problematic packages (e.g., os or   sys), which could lead to issues in execution.
    \item \textbf{Quality Evaluation:} We further evaluate the remaining functions using the StarCoder2-15B as a classifier to filter out examples with bad documentation or low-quality code.
    \item \textbf{Data Decontamination:} Finally, we employ an n-gram filtering technique to remove any functions that contain solutions or prompts from the benchmarks used in this work.
\end{itemize}

\section{Task Category}
\label{task_category}
Following the Magicoder \cite{DBLP:conf/icml/0003W0D024}, we use the following ten task categories for classifying code snippets: "Algorithmic and Data Structure Problems", "Mathematical and Computational Problems", "Database and SQL Problems", "System Design and Architecture Problems", "Security and Cryptography Problems", "Performance Optimization Problems", "Web Problems", "Domain Specific Problems", "User Interface and Application Design Problems", and "Data Science and Machine Learning Problems".

\section{Implementation Details}
\label{imple_detail}
\paragraph{Multi-Aspect Scorer.}
We sample 2.5K code instruction data from Llama3.1-8B-Ins, Qwen2.5-Coder-7B-Ins, Qwen2.5-Coder-14B-Ins, and the evol-codealpaca-v1 dataset \cite{DBLP:conf/iclr/LuoX0SGHT0LJ24}, respectively. Using the prompt in Appendix \ref{scorer}, we distill scoring results from GPT-4o-20240806 from $Z=10$ aspects and train Llama3.1-8B-Base for 3 epochs with a learning rate of $1 \times 10^{-5}$ and a global batch size of 64, obtaining the multi-aspect scorer. During inference, we set the temperature to 0. To derive the weight vector $\mathbf{w}$, we conduct $K=20$ experiments and evaluate the results on LiveCodeBench (202410-202501).

\paragraph{Reference Model.}
We train Llama3.1-8B-Base as the reference model on 10K  GPT-4o-20240806 data ($\mathcal{D}_{{p}}$) for 3 epochs with a learning rate of $2 \times 10^{-5}$ and a global batch size of 32. For LoRA configurations, we set $\text{lora\_r} = 128$, $\text{lora\_alpha} = 512$, and apply LoRA to the target modules: $\text{q\_proj}$, $\text{k\_proj}$, $\text{v\_proj}$, and $\text{o\_proj}$. We further investigate the impact of different reference models on data selection in Appendix~\ref{appendix:ref_model}.

\paragraph{Small-Scale Data synthesizer.}
We train Llama3.1-8B-Ins, Qwen2.5-Coder-7B-Ins, and Qwen2.5-Coder-14B-Ins as data synthesizers. Each model is first trained on 10K GPT-4o-20240806 data ($\mathcal{D}_{{p}}$) before undergoing two iterations of bootstrapping. In each iteration, we sample $N=3$ data synthesis samples from $M=5$ different checkpoints, respectively. The first iteration trains on 20K self-distilled samples, while the second iteration uses 40K. Each training runs for 3 epochs with a learning rate of $1 \times 10^{-5}$ and a batch size of 128. During inference, we set the temperature to 0.2.

\paragraph{SCoder.}
To maintain consistency with the baselines, we use DeepSeek-Coder-6.7B-Base as the base model and distill 60K code instruction samples from each of the three bootstrapped small-scale synthesizers. For a fair comparison, we also incorporate the evol-codealpaca-v1 dataset, an open-source Evol-Instruct implementation with approximately 110K data, widely used in baselines such as WizardCoder-GPT-4, WaveCoder-Ultra, MagicoderS, AlchemistCoder, and InverseCoder.  The training data size comparison across different models is presented in Table \ref{data_used}. 

To obtain SCoder, we first fine-tune DeepSeek-Coder-6.7B-Base on the 110K evol-codealpaca-v1 data for 2 epochs with an initial learning rate of $5 \times 10^{-5}$ and a global batch size of 512. We then further fine-tune it on the 60K small model-generated data for 3 epochs with an initial learning rate of $1 \times 10^{-5}$ and a batch size of 64. Both phases of training utilize a linear learning rate scheduler with a 0.05 warmup ratio and the AdamW optimizer. Training is conducted on 16 A100-80G GPUs.

\section{Prompts}
\label{magicoder_prompt}
\label{scorer}
\begin{figure*}[t!] 
    \centering 
    \includegraphics[width=1 \textwidth]{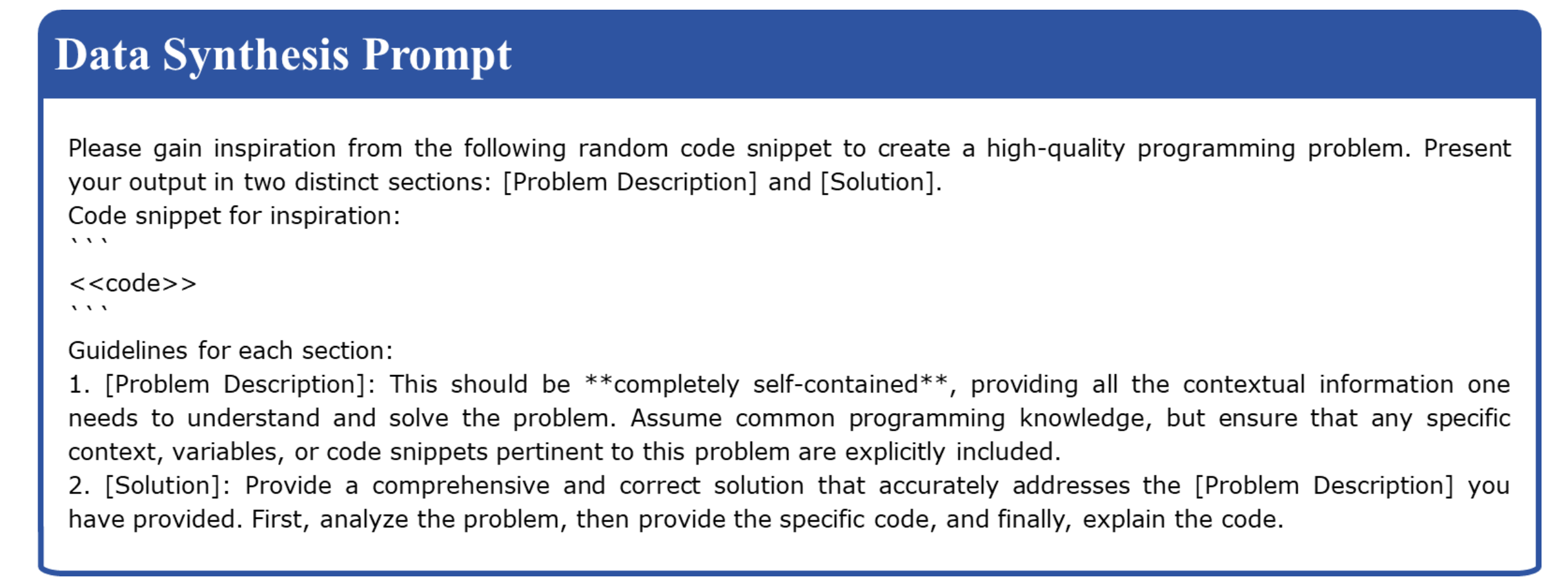} 
    \caption{Data Synthesis Prompt.} 
    \label{fig_6} 
\end{figure*}

\begin{figure*}[t!] 
    \centering 
    \includegraphics[width=1 \textwidth]{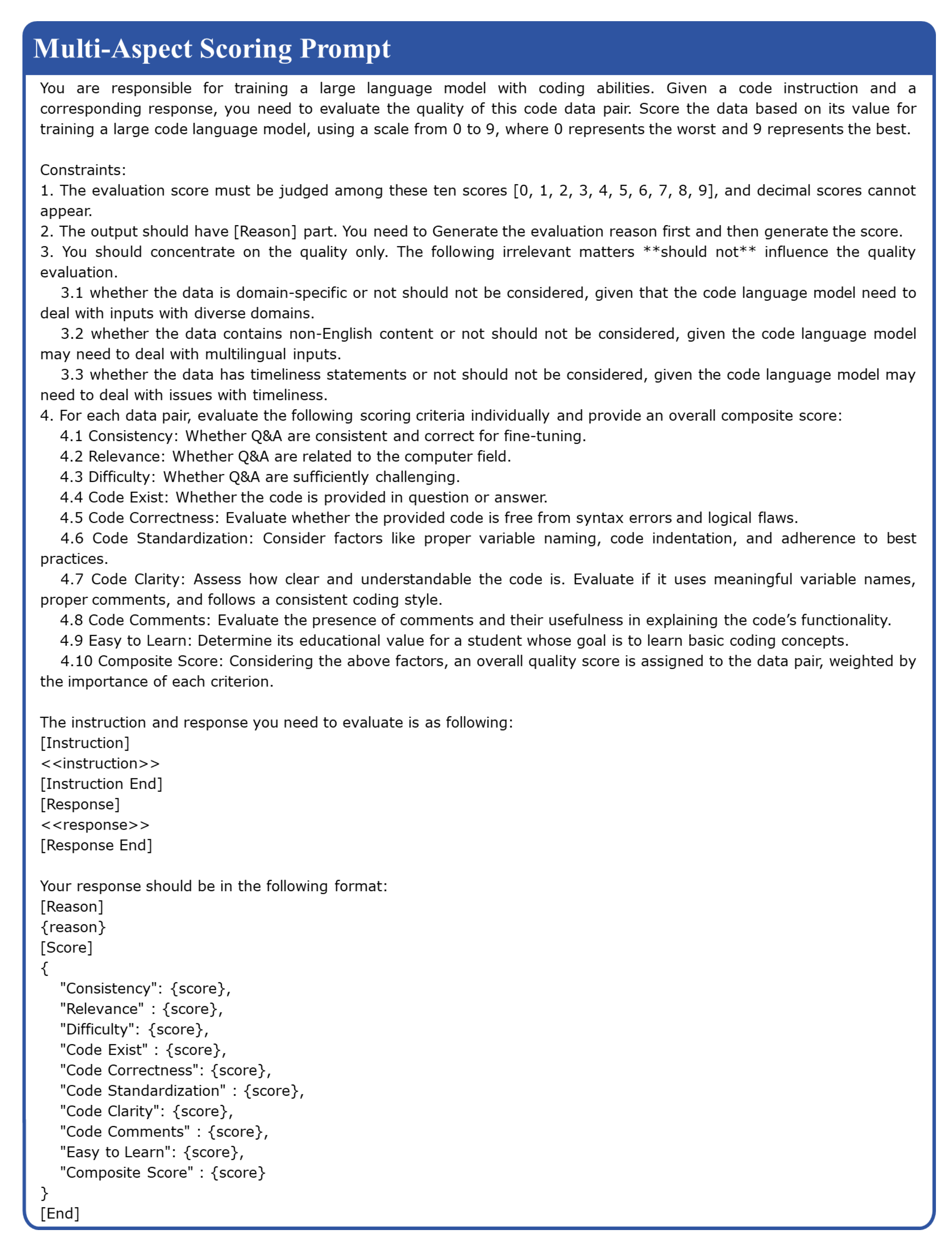} 
    \caption{Multi-Aspect Scoring Prompt.} 
    \label{fig_7} 
\end{figure*}

The  data synthesis prompt is inspired by ~\citet{DBLP:conf/icml/0003W0D024} and is shown in Figure \ref{fig_6}. The multi-aspect scoring prompt is inspired by ~\citet{DBLP:journals/corr/abs-2409-12186} and is shown in Figure \ref{fig_7}.

\section{Theoretical Analysis of Iterative Self-Distillation}
\label{sec:theoretical_analysis}

In this section, we provide a rigorous theoretical analysis of the  iterative self-distillation framework from two perspectives: convergence behavior and its interpretation in terms of Nash equilibrium and the exploration-exploitation trade-off.

\subsection{Problem Setup}

Let \( (\mathcal{M}, \| \cdot \|) \) be a complete metric space representing the space of model parameters. Let \( M_0 \in \mathcal{M} \) be a fixed initial model. Define the data generation process as a mapping \( \mathcal{G}: \mathcal{M} \rightarrow \mathcal{P} \), where \( \mathcal{P} \) denotes the space of data distributions. The training operator is defined as \( \mathcal{T}: \mathcal{M} \times \mathcal{P} \rightarrow \mathcal{M} \), mapping a model and a dataset to an updated model.

At each self-distillation iteration \( i \), the process proceeds as follows:

\begin{align}
D_i &= \mathcal{G}(M_i), \label{five equ} \\
M_{i+1} &= \mathcal{T}(M_0, D_i). \label{six equ}
\end{align}
where \( D_i \) is the data generated by model \( M_i \), and each new model \( M_{i+1} \) is trained from scratch using the fixed initialization \( M_0 \) and dataset \( D_i \).
\subsection{Convergence Analysis}

We analyze the convergence behavior of the model sequence \( \{M_i\} \) by examining the composed operator \( \Phi(M) = \mathcal{T}(M_0, \mathcal{G}(M)) \), which encapsulates the entire update process at each iteration of self-distillation. This operator provides a clear description of how the model \( M \) evolves after one iteration of self-distillation, starting from the fixed model \( M_0 \).

\paragraph{Assumptions:}
We impose the following assumptions:

\begin{itemize}
    \item \textbf{(A1)} \textit{Training Lipschitz Continuity:} There exists \( L_T > 0 \) such that for all \( D, D' \in \mathcal{P} \), the training process satisfies:
    \begin{equation}
    \| \mathcal{T}(M_0, D) - \mathcal{T}(M_0, D') \| \leq L_T \| D - D' \|.
    \end{equation}
    
    \item \textbf{(A2)} \textit{Data Generation Lipschitz Continuity:} There exists \( L_G > 0 \) such that for all \( M, M' \in \mathcal{M} \), the data generation process satisfies:
    \begin{equation}
    \| \mathcal{G}(M) - \mathcal{G}(M') \| \leq L_G \| M - M' \|.
    \end{equation}

    \item \textbf{(A3)} \textit{Contraction Condition:} The product of the Lipschitz constants satisfies:
    \begin{equation}
    L_T L_G < 1.
    \end{equation}
\end{itemize}

Under assumptions (A1) and (A2), we can establish the following lemma: The composed operator \( \Phi(M) = \mathcal{T}(M_0, \mathcal{G}(M)) \) is Lipschitz continuous with a constant of \( L_T L_G \). Specifically, for any two models \( M \) and \( M' \), we have the following inequality:

\begin{equation}
\begin{aligned}
&\| \Phi(M) - \Phi(M') \| 
\\
=& \| \mathcal{T}(M_0, \mathcal{G}(M)) - \mathcal{T}(M_0, \mathcal{G}(M')) \| \\
\leq &L_T \| \mathcal{G}(M) - \mathcal{G}(M') \| \\
\leq &L_T L_G \| M - M' \|.
\end{aligned}
\end{equation}

Now, we impose the contraction condition (A3), which ensures that \( \Phi \) is a contraction mapping. Since \( L_T L_G < 1 \), we can apply Banach's Fixed-Point Theorem to guarantee the existence of a unique fixed point \( M^* \in \mathcal{M} \) such that \( M^* = \Phi(M^*) \). Given this, we can analyze the convergence of the model sequence \( \{ M_i \} \), where \( M_{i+1} = \Phi(M_i) \). For any \( i \geq 0 \), the distance between \( M_{i+1} \) and \( M^* \) is given by:
\begin{equation}
    \begin{aligned}
\| M_{i+1} - M^* \| &= \| \Phi(M_i) - \Phi(M^*) \| \\
&\leq L_T L_G \| M_i - M^* \|.
\end{aligned}
\end{equation}
By recursively applying this inequality, we obtain:
\begin{equation}
    \begin{aligned}
\| M_{i+1} - M^* \| \leq (L_T L_G)^i \| M_0 - M^* \|.
\end{aligned}
\end{equation}
Since \( L_T L_G < 1 \), the factor \( (L_T L_G)^i \) decays exponentially, and thus the sequence \( \{ M_i \} \) converges to \( M^* \) at a linear rate.

Therefore, under the assumptions of Lipschitz continuity of both the training and data generation processes, and the contraction condition, the model sequence converges to a unique fixed point \( M^* \), with linear convergence determined by the product of the Lipschitz constants \( L_T L_G \).

\subsection{Nash Equilibrium Interpretation}

Beyond convergence, the fixed point \( M^* \) of the self-distillation process can also be interpreted through a game-theoretic lens as a \textit{Nash equilibrium}.

Consider each iteration of self-distillation as a two-player interaction:
\begin{itemize}
    \item \textbf{Teacher:} A model \( M \in \mathcal{M} \) that generates synthetic data via \( \mathcal{G}(M) \).
    \item \textbf{Student:} A fixed model \( M_0 \) that is retrained on the teacher’s generated data via \( \mathcal{T}(M_0, \mathcal{G}(M)) \).
\end{itemize}

The process evolves according to the update rule in Equation \ref{five equ} and \ref{six equ}
where the teacher at iteration \( i \) is \( M_i \), and the student is always initialized as \( M_0 \). The student updates its parameters based on the synthetic data provided by the teacher, effectively defining a best-response map from the teacher’s strategy to a new model.

At convergence, the fixed point \( M^* \) satisfies:
\begin{equation}
    M^* = \mathcal{T}(M_0, \mathcal{G}(M^*)),
\end{equation}
which indicates that when the teacher generates data using \( M^* \), retraining the student \( M_0 \) on that data simply reproduces the same model \( M^* \). Thus, neither the teacher nor the student can unilaterally change their behavior to improve the outcome, satisfying the condition for a Nash equilibrium.

This perspective emphasizes that  iterative self-distillation converges to a stable teacher–student pair, where the synthetic data and the resulting trained model are mutually consistent.

\subsection{Exploration–Exploitation Trade-off}

The iterative nature of self-distillation inherently embeds an exploration–exploitation mechanism.

\begin{itemize}
    \item \textbf{Exploration:} In each iteration \( i \), the teacher model \( M_i \) generates a new dataset \( D_i = \mathcal{G}(M_i) \), which may differ significantly from previous iterations. This promotes exploration of new data distributions, especially in the early stages when \( M_i \) is far from convergence.

    \item \textbf{Exploitation:} At every iteration, the student model is always retrained from the fixed initialization \( M_0 \). This exploits prior knowledge encoded in \( M_0 \), focusing learning on the current data \( D_i \).
\end{itemize}

As training progresses, the diversity of generated data typically decreases, and the model converges to a stable state \( M^* \). In this sense, the process naturally transitions from high-entropy exploration to low-entropy exploitation. This dynamic provides a theoretical rationale for the empirical success of  iterative self-distillation.

\subsection{Discussion}

\begin{table}[t!]
\centering
\begin{tabular}{lcc}
\toprule
\textbf{Data Synthesizer} & \textbf{HE} & \textbf{LCB-V4-Full} \\
\midrule
Llama3.1-8B      & 60.4               & 16.5                            \\
\ \ \ \ +2 iter                   & 67.4               & 17.8                            \\
\ \ \ \ +3 iter                   & 67.2               & 17.9                            \\
Qwen2.5-Coder-7B & 61.6               & 17.0                            \\
\ \ \ \ +2 iter                   & 68.9               & 18.9                            \\
\ \ \ \ +3 iter                   & 69.1               & 18.8                            \\
\bottomrule
\end{tabular}
\caption{Finetuning performance of DeepSeek-Coder-6.7B-Base on 40K data synthesized by different synthesizers.}
\label{tab:iter_performance}
\end{table}

Although the convergence and equilibrium are guaranteed under idealized assumptions (e.g., Lipschitz continuity, contraction property), in practical scenarios with non-convex models and imperfect optimization, strict convergence is not guaranteed. However, our empirical results suggest that the self-distillation process stabilizes in practice and leads to consistently improved model performance, as shown in Table \ref{bootstrap_table}.

Furthermore, we extended the self-distillation process to three iterations. In the third iteration, we used 40K self-distilled samples for training. As shown in Table~\ref{tab:iter_performance}, the performance of the synthesizers becomes stable when the number of self-distillation iterations reaches two or more, indicating that additional iterations yield diminishing returns while maintaining strong generation quality.

\section{Cost Efficiency of Our Method}
\label{appendix:cost}

In this section, we detail the cost advantages of our proposed approach, which relies on training a lightweight data synthesizer rather than directly distilling a proprietary large language model (LLM). Our method significantly reduces reliance on expensive LLM queries, improving both efficiency and accessibility.

Specifically, we use only 10K proprietary LLM samples during the initial bootstrapping phase. This is a substantial reduction compared to prior works, which typically require 150K–200K proprietary samples, as shown in Table~\ref{data_used}. By contrast, once the bootstrapped synthesizer is trained, we can generate high-quality instruction data at scale without further calls to proprietary models.

The main computational cost of our method lies in fully fine-tuning the data synthesizer. In comparison, model inference (for sampling and multi-aspect scoring) and gradient similarity calculations are relatively lightweight. For instance, constructing the gradient library for each iteration takes approximately 3 hours on a single NVIDIA A100 80GB GPU.

Taking Qwen2.5-Coder-7B-Instruct as an example, we fine-tuned on 110K self-distilled samples throughout the entire bootstrap process, which took around 6.5 hours on 8× A100 80GB GPUs. Based on Google Cloud's official pricing\footnote{\url{https://cloud.google.com/products/calculator}}, the total cost is estimated to be only \$263.58. In contrast, using proprietary model APIs such as the GPT-4o-20240806 API for instruction synthesis incurs significantly higher costs; given average input/output lengths of 253 and 752 tokens respectively (as statistically measured from 10K distilled samples from the proprietary model), the same budget would only allow for generating approximately 30K samples. This highlights the efficiency of our approach: once trained, the synthesizer enables large-scale data generation at a fraction of the cost.

\section{Influence of Different Reference Models}
\label{appendix:ref_model}

In our main experiments, we primarily used Llama3.1-8B-Base as the reference model to compute gradient-based influence scores for guiding data selection. To assess whether the choice of reference model significantly impacted the outcome, we conducted additional experiments using different reference models while keeping the data synthesizer (Qwen2.5-Coder-7B-Instruct) and the target model (DeepSeek-Coder-6.7B-Base) unchanged.

We trained the data synthesizers for 2 iterations and used them to generate 40K code instruction data for training the target model. As shown in Table~\ref{tab:ref_model}, the performance variations across different reference models are relatively small. This indicates that our method is stable and largely insensitive to the specific scale or version of the reference model, further validating its robustness and practicality.

\begin{table}[t!]
\centering
\begin{tabular}{lcc}
\toprule
\textbf{Reference Model} & \textbf{HE} & \textbf{LCB-V4-Full} \\
\midrule
Llama3.1-8B & 68.9 & 18.9 \\
Llama2-7B & 68.4 & 18.5 \\
Llama2-13B & 69.2 & 18.8 \\
\bottomrule
\end{tabular}
\caption{Performance of the target model (DeepSeek-Coder-6.7B-Base) using different reference models for influence estimation. The data synthesizer is fixed to Qwen2.5-Coder-7B-Instruct.}
\label{tab:ref_model}
\end{table}

\section{Data validity on more target models}
\label{moretargetmodel}
To further demonstrate the generalization capability of the data synthesized by the small synthesizers, we additionally selected Llama3.1-8B-Base and Qwen2.5-Coder-7B-Base as target models. Following the settings described in Appendix \ref{imple_detail}, we set the bootstrapped Qwen2.5-Coder-14B-Ins as the synthesizer and trained SCoder-Q14-Llama-8B and SCoder-Q14-Qwen-7B respectively. As shown in Table \ref{tabmoretarget}, SCoder achieves significant improvements over the corresponding instruction models across the majority of evaluation metrics. Considering that Qwen2.5-Coder-7B-Ins was trained on millions of instruction data while we only used 60K data generated by the small synthesizer, this still demonstrates the effectiveness of our approach.

\begin{table}[t!]
\centering
\begin{tabular}{lcccc}
\toprule
\textbf{Models} & \multicolumn{2}{c}{\textbf{HE}} & \multicolumn{2}{c}{\textbf{LCB-V4}}\\
& Base & Plus & Full & Easy
\\
\midrule
Llama3.1-8B-Ins & 65.9 & 57.9 & 18.0& 46.7\\
SCoder-Q14-Llama-8B & 70.1 & 64.6 & 19.1 & 48.3 \\
Qwen2.5-Coder-7B-Ins & 88.4\textsuperscript{†} & 84.1\textsuperscript{†} & 24.7 & 36.6 \\
SCoder-Q14-Qwen-7B & 85.8 & 80.0 & 29.4 & 65.2 \\

\bottomrule
\end{tabular}
\caption{Training results of different target models. † denotes results from the official technical report.}
\label{tabmoretarget}
\end{table}

\end{document}